\documentclass[sigconf]{acmart}
\AtBeginDocument{%
  }

\usepackage{multirow}
\usepackage{amsthm}
\usepackage{graphicx}
\usepackage{subcaption}
\usepackage{pgfplots}
\usetikzlibrary{matrix}

\usepgfplotslibrary{statistics}
\usepackage{algorithm}
\usepackage{algpseudocode}
\usepackage{hyperref}
\usepackage{bm}

\begin{document}

% Anomaly Detection from Dynamic Graph with Normality Distribution Shift
\title{Robust Anomaly Detection Under Normality Distribution Shift \\ in Dynamic Graphs}

% \if0
\author{Xiaoyang Xu}
\email{xu.xiaoyang.42r@st.kyoto-u.ac.jp}
\affiliation{%
  \institution{Kyoto University}
  \city{Kyoto}
  \country{Japan}
}

\author{Xiaofeng Lin}
\email{lxf@ml.ist.i.kyoto-u.ac.jp}
\affiliation{%
  \institution{Kyoto University}
  \city{Kyoto}
  \country{Japan}
}

\author{Koh Takeuchi}
\email{takeuchi@i.kyoto-u.ac.jp}
\affiliation{%
  \institution{Kyoto University}
  \city{Kyoto}
  \country{Japan}
}

\author{Kyohei Atarashi}
\email{atarashi@i.kyoto-u.ac.jp}
\affiliation{%
  \institution{Kyoto University}
  \city{Kyoto}
  \country{Japan}
}

\author{Hisashi Kashima}
\email{kashima@i.kyoto-u.ac.jp}
\affiliation{%
  \institution{Kyoto University}
  \city{Kyoto}
  \country{Japan}
}

\renewcommand{\shortauthors}{Xu et al.}
% \fi

% \author{Anonymous Author(s)}
% \email{}
% \affiliation{%
%   \institution{}
%   \city{}
%   \country{}
% }
% \renewcommand{\shortauthors}{Anonymous Author(s)}

\begin{abstract}
Anomaly detection in dynamic graphs is a critical task with broad real-world applications, including social networks, e-commerce, and cybersecurity. Most existing methods assume that normal patterns remain stable over time; however, this assumption often fails in practice due to the phenomenon we refer to as normality distribution shift (NDS), where normal behaviors evolve over time. Ignoring NDS can lead models to misclassify shifted normal instances as anomalies, degrading detection performance. To tackle this issue, we propose WhENDS, a novel unsupervised anomaly detection method that aligns normal edge embeddings across time by estimating distributional statistics and applying whitening transformations. Extensive experiments on four widely-used dynamic graph datasets show that WhENDS consistently outperforms nine strong baselines, achieving state-of-the-art results and underscoring the importance of addressing NDS in dynamic graph anomaly detection. The codes are available at https://anonymous.4open.science/r/WhENDS.
\end{abstract}





\maketitle

\section{Introduction}
Anomaly detection, which aims to identify patterns in data that deviate from expected norms~\cite{anomalydetectionsurvey2009}, is a fundamental task with critical applications across various domains. For instance, in social networks, interactions among benign users represent normal behavior, while activities such as fraud, cyberbullying, phishing, and rumor propagation are considered anomalies. Accurately detecting such anomalies is vital for ensuring system security and maintaining platform integrity.

In many real-world scenarios, data can be naturally modeled as dynamic graphs—graph-structured data whose topology and features evolve over time. Typical examples include social networks~\cite{socialnetwork2006,socialnetwork2019}, financial systems~\cite{financialnetwork2019,financialnetwork2021}, and traffic networks~\cite{trafficnetwork2023}. As a result, anomaly detection in dynamic graphs has emerged as an important and actively studied problem. Due to the scarcity of labeled anomalies, most existing dynamic graph anomaly detection methods~\cite{netwalk2019,taddy2021,rustgraph2024} adopt an unsupervised approach: they learn the distribution of normal patterns from training data and detect anomalies as deviations from this distribution during testing.

%######################################
\begin{figure}
    \centering
    \includegraphics[height=4cm,width=8cm]{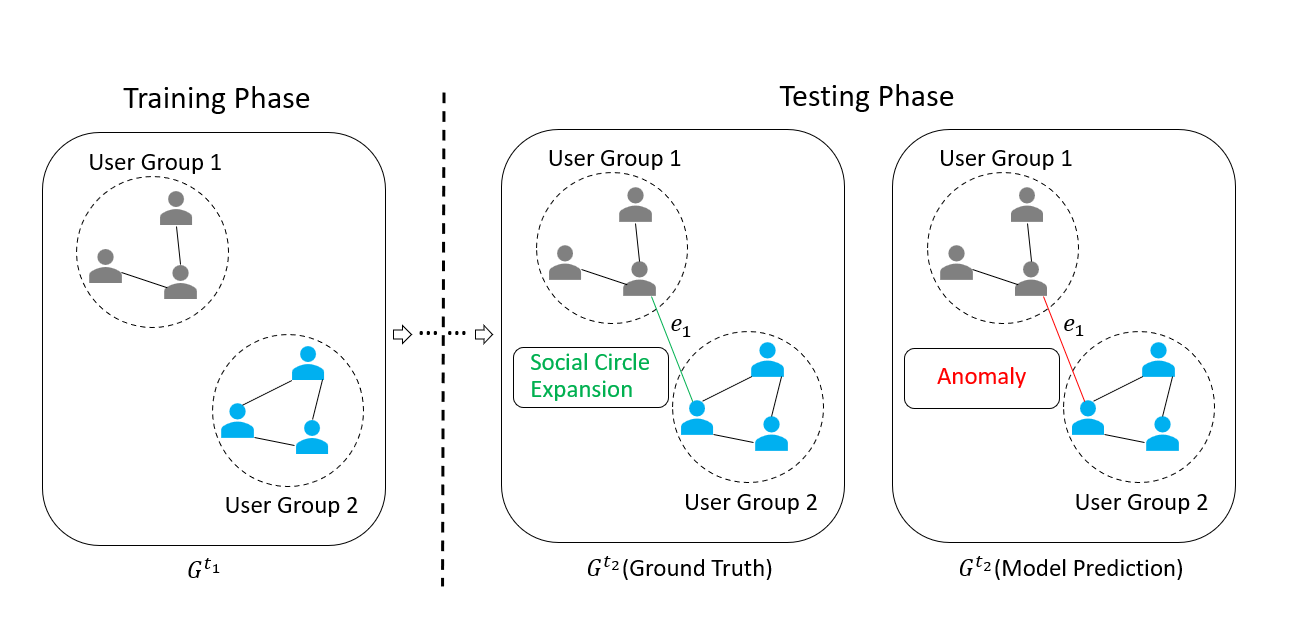}
    \caption{A toy example illustrating how normality distribution shift (NDS) affects model performance. During training, two groups of users have no inter-group connections. In the test phase, a new edge $e_1$ (in green) emerges due to natural behavioral evolution, such as the expansion of social circles. Although $e_1$ should be considered normal, it deviates from the normal pattern learned during training and is thus mistakenly classified as an anomaly (in red).}
    \label{NDS}
\end{figure}
%#########################################
% \begin{figure}
%     \centering
%     \includegraphics[height=4cm,width=8cm]{intro4.png}
%     \caption{A toy example illustrating how normality distribution shift (NDS) affects model performance. During training, two groups of normal users have no inter-group connections. In the test phase, two new edges ($e_1$ and $e_2$) deviate from this learned pattern. Edge $e_1$ results from natural behavioral evolution (e.g., expansion of social circles), while $e_2$ is introduced by a newly joined malicious user. Although $e_1$ represents a benign behavior, it deviates from the training distribution similarly to $e_2$. As a result, the model incorrectly classifies both as anomalies (in red), which lead to degraded detection performance.}
%     \label{NDS}
% \end{figure}

Although existing approaches have shown promising performance, they typically assume that normal patterns remain stable over time. In reality, however, normal behavior may evolve—a phenomenon we refer to as \emph{normality distribution shift (NDS)}. For example, in social networks, benign user behavior may vary due to seasonal trends, emerging events, or natural expansion of social circles. As illustrated in Figure~\ref{NDS}, the presence of NDS means that even normal samples may deviate from the patterns learned during training, potentially leading models to misclassify them as anomalies.

To tackle this challenge, we propose a novel anomaly detection method for dynamic graphs, called \underline{\textbf{Wh}}itening of \underline{\textbf{E}}dge distribution for \underline{\textbf{N}}ormality \underline{\textbf{D}}istribution \underline{\textbf{S}}hift (\textbf{WhENDS}). WhENDS estimates the distributional statistics of normal edge embeddings at each timestamp and uses them to perform whitening. This alignment transforms normal edge embeddings across time into a common standard Gaussian distribution, thereby mitigating the impact of NDS.

We summarize the main contributions of this work as follows:
\begin{itemize}
    \item We identify and formalize the challenge of normality distribution shift (NDS) in dynamic graph anomaly detection, and empirically demonstrate its significant impact on model performance.
    \item We propose \textbf{WhENDS}, a novel unsupervised anomaly detection method that mitigates NDS by estimating and whitening the distribution of normal edge embeddings over time.
    \item Extensive experiments on benchmark datasets show that WhENDS outperforms state-of-the-art methods and remains robust under various levels of NDS.
\end{itemize}

\section{Related Work}
\subsection{Anomaly Detection in Dynamic Graphs}
% \vspace{1ex}
% \noindent\textbf{Anomaly Detection in Dynamic Graphs.} 
As an important task in dynamic graphs, anomaly detection has attracted significant research attention. Early methods primarily relied on traditional machine learning techniques. For example, GOutlier~\cite{goutlier2011} identifies anomalies by analyzing structural connectivity patterns, and CM-Sketch~\cite{cmsketch2016} leverages approximated anomaly metrics derived from empirical statistics to detect abnormal edges.

In recent years, the strong performance of deep learning has made it the dominant paradigm for dynamic graph anomaly detection. NetWalk~\cite{netwalk2019} adopts a reconstruction-based strategy by training an autoencoder on random walk sequences to generate node and edge embeddings, and evaluates anomaly scores through clustering these embeddings. AddGraph~\cite{addgraph2019} leverages GCN~\cite{gcn2017} to encode graph information and employs an attention mechanism along with GRU~\cite{gru2014} to capture both short-term and long-term temporal dependencies. StrGNN~\cite{strgnn2021} encodes h-hop enclosing subgraphs using GCN and aggregates temporal information with GRU. TADDY~\cite{taddy2021} utilizes edge-based substructure sampling and a Transformer~\cite{transformer2017} to obtain spatial-temporal encodings of dynamic graphs. RustGraph~\cite{rustgraph2024} integrates GNNs with GRUs to aggregate temporal information and introduces VGAE~\cite{vage2016} and graph contrastive learning to better encode dynamic graphs.

Despite their success, existing dynamic graph anomaly detection methods generally assume that the distribution of normal patterns remains stationary over time. They do not explicitly consider the possibility of NDS, where the normal pattern can change over time. As a result, these methods may misclassify shifted normal samples as anomalies, leading to performance degradation under NDS.

\subsection{Graph Out-Of-Distribution~(OOD) Generalization}
% \vspace{1ex}
% \noindent\textbf{Graph Out-Of-Distribution~(OOD) Generalization.} 
To address distribution shift in graph data, numerous Graph OOD generalization methods have been proposed. These methods can be broadly categorized into three groups~\cite{graphoodsurvey2025}: data-augmentation-based methods~\cite{gaug2023,ncla2023,grand2021,graphcl2021}, generalizable model-based methods~\cite{disgnn2019,oodgnn2023}, and learning strategy-based methods~\cite{stablegl2023,pretraininggnn}. However, most of these studies focus on static graphs and fail to capture the crucial temporal information inherent in dynamic graphs, making them unsuitable for direct application to dynamic settings.

In an effort to handle distribution shift in dynamic graphs, several recent works have explored OOD generalization methods specifically for dynamic graphs~\cite{dygood1,dygood2,dygood3}. Nevertheless, these approaches typically rely on large amounts of labeled data for training, which limits their applicability to anomaly detection in dynamic graphs, where labeled anomalies are scarce or unavailable.

Despite these efforts in generalizing the OOD of graphs, the aforementioned limitations prevent existing methods from being directly applied to anomaly detection in dynamic graphs to address the problem of NDS, which motivates the design of our proposed method.

\begin{figure*}
    \centering
    \includegraphics[trim=20 100 20 100, clip, height=8cm, width=16cm]{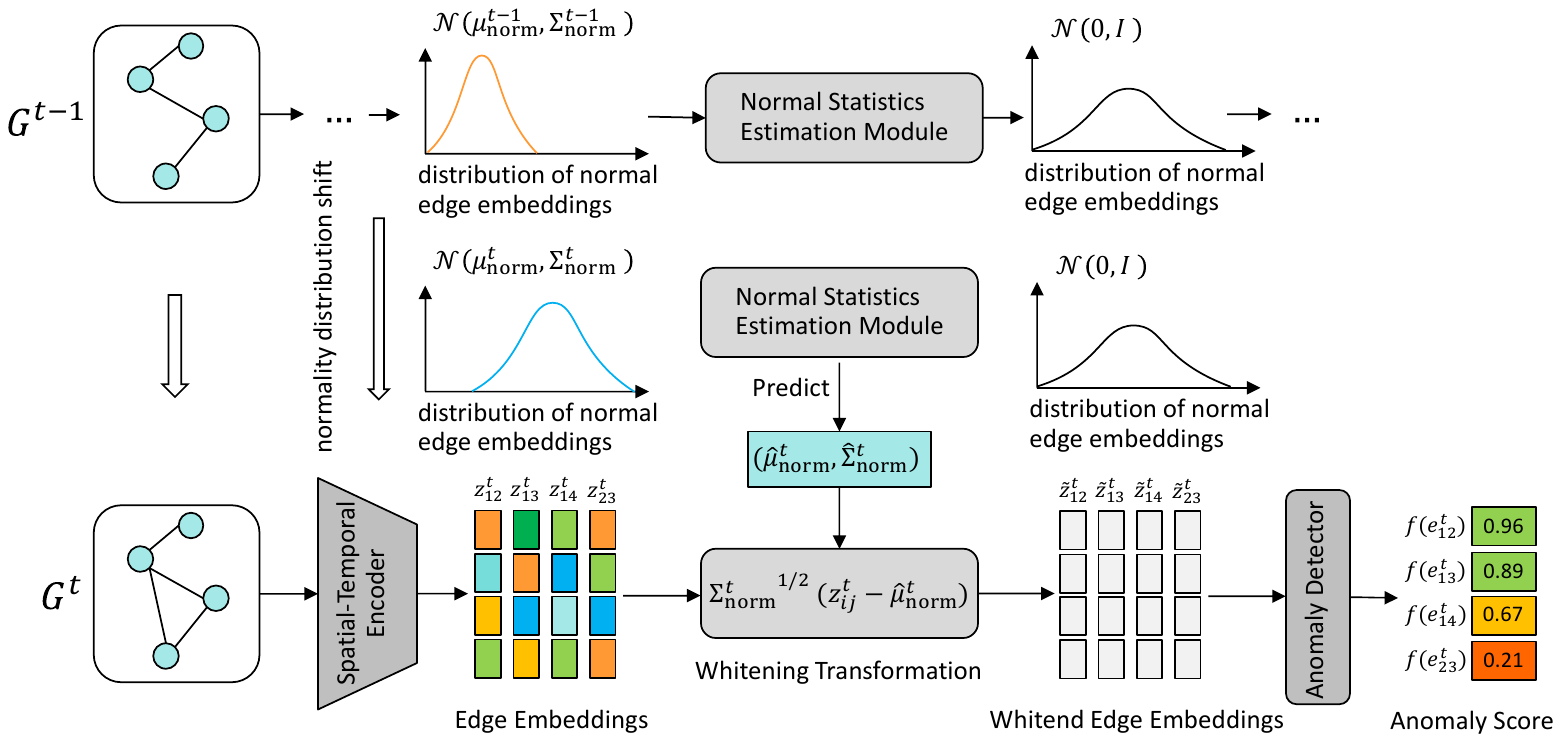}
\caption{Overview of WhENDS. The framework consists of three main modules: a Spatial-Temporal Encoder, a Normal Statistics Estimation Module (NSEM), and an Anomaly Detector. At each timestamp, the input graph snapshot is processed by the Spatial-Temporal Encoder to obtain edge embeddings, which are then whitened using predicted statistics from the NSEM. The whitened embeddings are passed to the Anomaly Detector to compute anomaly scores for each edge. By aligning normal edge embeddings at all timestamps to a standard Gaussian distribution, the NSEM mitigates the effects of normality distribution shift.}
    \label{WhENDS}
\end{figure*}

\section{Problem Formulation}
Let $\mathbb{G} = \{\mathcal{G}^{1}, \mathcal{G}^{2}, \dots, \mathcal{G}^{T}\}$ be a dynamic graph with maximum timestamp $T$, where each snapshot $\mathcal{G}^{t}$ at timestamp $t$ is defined as $\mathcal{G}^{t} = \{\mathcal{V}^{t}, \mathcal{E}^{t}\}$, with $\mathcal{V}^{t}$ denoting the node set and $\mathcal{E}^{t}$ the edge set. For attributed graphs, we use $\{\boldsymbol{A}^{t}, \boldsymbol{X}^{t}\}$, where $\boldsymbol{A}^{t} \in \mathbb{R}^{N^{t} \times N^{t}}$ is the adjacency matrix such that $\boldsymbol{A}^{t}_{ij} = 1$ if an edge $e^{t}_{ij}$ exists between nodes $v^{t}_{i}$ and $v^{t}_{j}$, and $0$ otherwise; $\boldsymbol{X}^{t} \in \mathbb{R}^{N^{t} \times d}$ denotes the node feature matrix, where $N^{t} = |\mathcal{V}^{t}|$ and $d$ is the feature dimension. Over time, both $\mathcal{V}^{t}$, $\mathcal{E}^{t}$, and node attributes may evolve.

To distinguish between normal and anomalous edges in each snapshot $\mathcal{G}^{t}$, we denote the sets of normal and anomalous edges as $\mathcal{E}^{t}_{\text{norm}}$ and $\mathcal{E}^{t}_{\text{ano}}$, respectively. We define \emph{normality distribution shift (NDS)} in a dynamic graph as a temporal change in the distribution of normal edges. Let $\mathcal{D}_{\text{norm}}^{t}$ represent the distribution of normal edges $e_{ij}^{t} \in \mathcal{E}^{t}_{\text{norm}}$ at timestamp $t$. Then, NDS occurs if $\mathcal{D}_{\text{norm}}^{t_1} \neq \mathcal{D}_{\text{norm}}^{t_2}$ for any $t_1 \neq t_2$.

Following prior work~\cite{addgraph2019, taddy2021, strgnn2021, rustgraph2024}, we adopt an unsupervised setting for anomaly detection in dynamic graphs. Specifically, the data in the training phase are assumed to be entirely normal and unlabeled; we use the first $K$ snapshots $\mathbb{G}_{\text{tr}} = \{\mathcal{G}^{1}, \dots, \mathcal{G}^{K}\}$ as the training set. During testing, we inject synthetic anomalous edges into the remaining snapshots $\mathbb{G}_{\text{te}} = \{\mathcal{G}^{K+1}, \dots, \mathcal{G}^{T}\}$ to evaluate model performance.

Our goal is to learn an anomaly detection function $f(\cdot)$ using only the training data. Given an edge $e_{ij}^{t}$, the model outputs an anomaly score $f(e_{ij}^{t}) \in [0, 1]$, representing the likelihood that the edge is anomalous. The model should assign low scores to normal edges and high scores to anomalous ones, and maintain robustness even under NDS.

% \begin{figure*}
%     \centering
%     \includegraphics[trim=20 100 20 100, clip, height=8cm, width=16cm]{WhENDS.pdf}
%     \caption{Overview of WhENDS. WhENDS consists of three main modules: a Spatial-Temporal Encoder, a Normal Statistics Estimation Module (NSEM), and an Anomaly Detector. At each timestamp, the input snapshot is encoded by the Spatial-Temporal Encoder to obtain edge embeddings, which are then whitened using the predicted statistics from the NSEM. The whitened embeddings are fed into the Anomaly Detector to compute anomaly scores for each edge. By aligning normal edge embeddings at all timestamps to a standard Gaussian distribution, the NSEM mitigates the effects of normality distribution shift. }
% \end{figure*}
\section{Proposed Method}
In this section, we present the components of our proposed method, \textbf{WhENDS}, as illustrated in Figure~\ref{WhENDS}. WhENDS is composed of three key modules:
(1) the \textbf{Spatial-Temporal Encoder}, which captures spatial-temporal patterns in dynamic graphs and generates edge embeddings;
(2) the \textbf{Normal Statistic Estimation Module (NSEM)}, which estimates the distributional statistics of normal edge embeddings and performs whitening to align them with a standard Gaussian distribution at each timestamp; and
(3) the \textbf{Anomaly Detector}, which assigns an anomaly score to each edge based on the whitened embeddings.

We begin by detailing the Normal Statistic Estimation Module, the core component of WhENDS responsible for mitigating the effects of NDS.

\subsection{Normal Statistics Estimation Module}
%%%%%%%%%%%%%%%%%%%%%%% CIKM Version%%%%%%%%%%%%%%%%%%%%%%%%%%%
% Our method is based on the following assumption:
% at any timestamp $t$, the embeddings of normal edges can be approximately modeled by a multivariate Gaussian distribution $\mathcal{N}(\boldsymbol{\mu}_{\text{norm}}^t, \boldsymbol{\Sigma}_{\text{norm}}^t)$.

%%%%%%%%%%%%%%%%%%%%%%% Revised Version %%%%%%%%%%%%%%%%%%%%%%%%%%%%
To model the distribution of normal edges and address the challenge of NDS, we make the following approximation: at any timestamp $t$, the embeddings of normal edges can be approximately modeled by a multivariate Gaussian distribution $\mathcal{N}(\boldsymbol{\mu}_{\text{norm}}^t, \boldsymbol{\Sigma}_{\text{norm}}^t)$. While this approximation may not perfectly capture all real-world edge distributions, it provides a principled and efficient basis for mitigating NDS in dynamic graphs. 
Under this approximation, if the parameters $(\boldsymbol{\mu}_{\text{norm}}^t, \boldsymbol{\Sigma}_{\text{norm}}^t)$ can be obtained, a whitening transformation can be applied to align the normal edge embeddings at each timestamp to a standard Gaussian distribution $\mathcal{N}(0, I)$, thus mitigating the impact of NDS.
%%%%%%%%%%%%%%%%%%%%%%%%%%%%%%%%%%%%%%%%%%%%%%%%%%%%%%%%%%%%%%%%%%%%%
\begin{figure}[htbp]
    \centering
    \includegraphics[trim=149 33 149 19, clip,height=7cm, width=7cm]{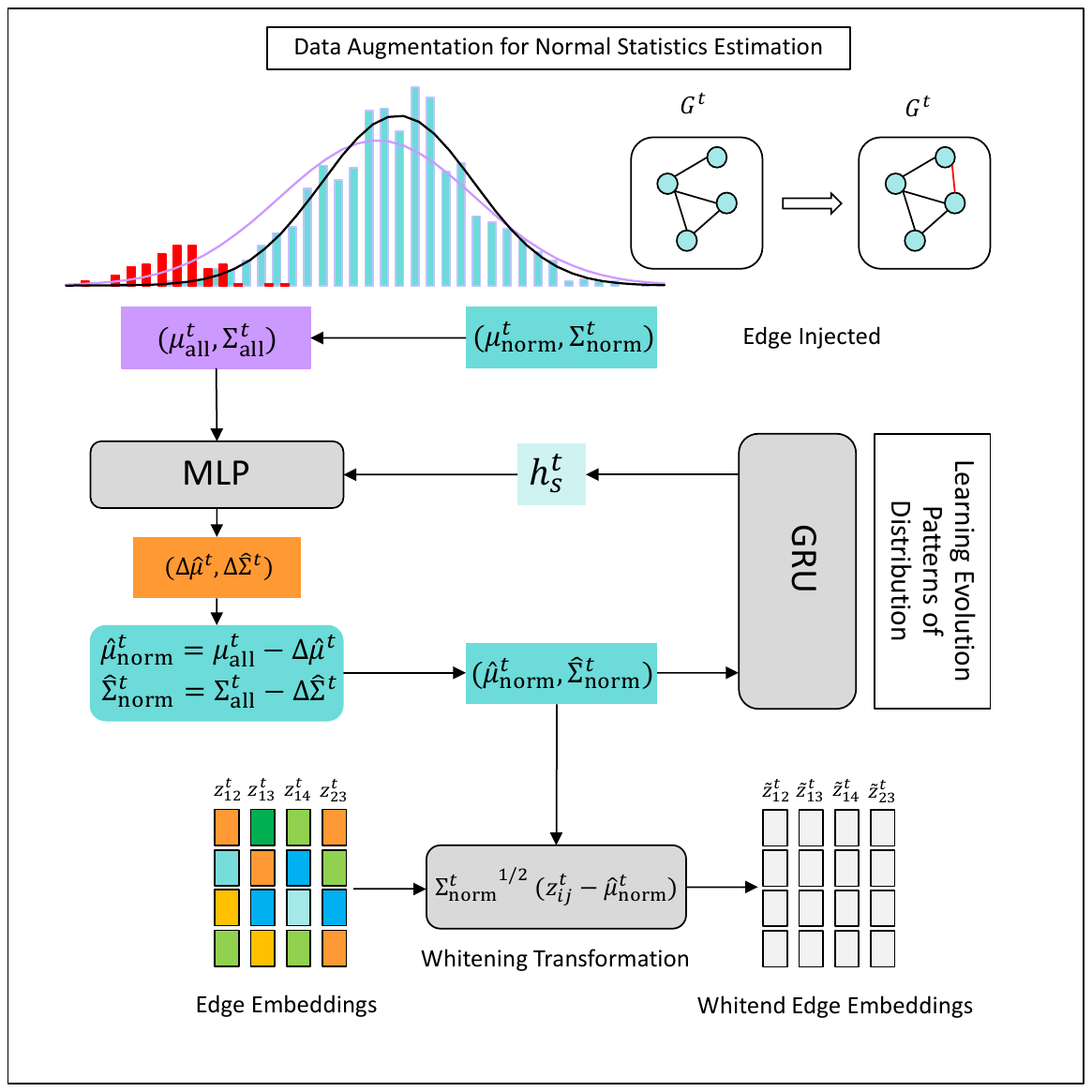}
\caption{Normal Statistics Estimation Module. The NSEM consists of an MLP and a GRU. Given the overall statistics as input, the MLP estimates the deviation from normal statistics and recovers the clean normal statistics, which are then used to whiten the edge embeddings at the current timestamp. The predicted normal statistics are also passed to the GRU to update its hidden state, allowing the model to capture temporal evolution and improve predictions at subsequent timestamps.}
    \label{NSEM}
\end{figure}

Since the true statistics of normal edge embeddings are not directly accessible in the testing phase, we design the Normal Statistics Estimation Module (NSEM) to estimate the statistics of normal edge embeddings $(\boldsymbol{\mu}_{\text{norm}}^t, \boldsymbol{\Sigma}_{\text{norm}}^t)$ at each timestamp $t$, as illustrated in Figure~\ref{NSEM}. NSEM leverages data augmentation to learn the deviation between the overall edge distribution and the normal edge distribution in the presence of anomalies. Additionally, a GRU is employed to model the evolving pattern of the distribution of normal edges. The estimated statistics are then used to perform whitening on the edge embeddings.

\vspace{1ex}
\noindent\textbf{Data Augmentation for Normal Statistics Estimation.} In the training phase, there is no anomalous edge present. Therefore, for each snapshot $\mathcal{G}^t$ in the training set, the distribution of all edges in $\mathcal{G}^t$ is equivalent to that of the normal edges at timestamp $t$, i.e.,
\begin{equation}
    \mathcal{D}_{\text{all}}^t = \mathcal{D}_{\text{norm}}^t, \quad \text{if} \quad t \leq K.
    \label{trdist}
\end{equation}
In contrast, in the testing phase, due to the presence of anomalous edges, the distribution over all edges in $G^t$ becomes
\begin{equation}
    \mathcal{D}_{\text{all}}^t = \mathcal{D}_{\text{norm}}^t \cup \mathcal{D}_{\text{ano}}^t, \quad \text{if} \quad t > K.
    \label{tedist}
\end{equation}
The presence of anomalies causes the overall edge embedding statistics (denoted as $(\boldsymbol{\mu}_{\text{all}}^t,\boldsymbol{\Sigma}_{\text{all}}^t)$) to deviate from the normal edge distribution. Denote $\Delta\boldsymbol{\mu}^t,\Delta\boldsymbol{\Sigma}^t$ as this deviation:
\begin{align}
    &\Delta\boldsymbol{\mu}^t=\boldsymbol{\mu}_{\text{all}}^t-\boldsymbol{\mu}_{\text{norm}}^t,\label{deltamu}\\
    &\Delta\boldsymbol{\Sigma}^t=\boldsymbol{\Sigma}_{\text{all}}^t-\boldsymbol{\Sigma}_{\text{norm}}^t.\label{deltasigma}
\end{align}
To solve this, a natural idea is to estimate and subtract this deviation $\Delta\boldsymbol{\mu}^t,\Delta\boldsymbol{\Sigma}^t$ in order to recover the normal statistics $(\boldsymbol{\mu}_{\text{norm}}^t, \boldsymbol{\Sigma}_{\text{norm}}^t)$ from the corrupted observations. To achieve this, we simulate the presence of anomalies via data augmentation on the training set. Specifically, for each training snapshot, we randomly inject $\alpha$ (sampled from 1\% to 10\%) of synthetic edges, which are treated as anomalous. We then use a multi-layer perceptron (MLP) to predict $(\Delta\boldsymbol{\mu}^t,\Delta\boldsymbol{\Sigma}^t)$ on the augmented snapshot:
\begin{equation}
    \Delta \hat{\boldsymbol{\mu}}^t,\Delta \hat{\boldsymbol{\Sigma}}^t=\text{MLP}(\boldsymbol{\mu}_{\text{all}}^t\oplus \text{Flatten}(\text{Chol}(\boldsymbol{\Sigma}_{\text{all}}^t))\oplus \boldsymbol{h}_{s}^{t-1}),\label{NSEMmlp}
\end{equation}
where $\oplus$ denotes the concatenation operator, $\text{Chol}(\cdot)$ denotes the Cholesky decomposition that extracts the lower triangular matrix, and ${\text{Flatten}}(\cdot)$ denotes the operation that flattens a matrix into a one-dimensional vector. With the predicted deviations $\Delta \hat{\boldsymbol{\mu}}^t$ and $\Delta \hat{\boldsymbol{\Sigma}}^t$, we recover the estimated statistics of normal edge embeddings as:
\begin{align}
    &\hat{\boldsymbol{\mu}}_{\text{norm}}^t=\boldsymbol{\mu}_{\text{all}}^t-\Delta \hat{\boldsymbol{\mu}}^t,\label{predmu}\\
    &\hat{\boldsymbol{\Sigma}}_{\text{norm}}^t=(\boldsymbol{\Sigma}_{\text{all}}^t-\Delta \hat{\boldsymbol{\Sigma}}^t)(\boldsymbol{\Sigma}_{\text{all}}^t-\Delta \hat{\boldsymbol{\Sigma}}^t)^\top.\label{predsigma}
\end{align}
In the above equations, $(\boldsymbol{\Sigma}_{\text{all}}^t-\Delta \hat{\boldsymbol{{\Sigma}}}^t)(\boldsymbol{\Sigma}_{\text{all}}^t-\Delta \hat{\boldsymbol{\Sigma}}^t)^\top$ is designed to ensure that $\hat{\boldsymbol{\Sigma}}_{\text{norm}}^t$ is positive definite, which guarantees the validity of the subsequent whitening transformation.

\vspace{1ex}
\noindent \textbf{Learning Evolution Patterns of Distribution.} 
%%%%%%%%%%%%%%%%%%%%%%%%%%%%%%%%% CIKM Version %%%%%%%%%%%%%%%%%%%%%%%%%%%%%%%%%
% Many prior studies have shown that evolving data distributions often follow certain temporal patterns~\cite{edg2022,edg2023,edg2024}. Capturing such patterns may improve the accuracy of the estimation of statistics. To this end, we employ a Gated Recurrent Unit (GRU)~\cite{gru2014} to model the evolution pattern of normal edge distributions and update the hidden states $h_{s}^t$ in Equation~\eqref{NSEMmlp}:
%%%%%%%%%%%%%%%%%%%%%%%%%%%%%%%%% Revised Version %%%%%%%%%%%%%%%%%%%%%%%%%%%%%%
In our design, NSEM is trained using augmented data in which random edges are inserted into the training snapshots to simulate anomalies. However, in practice, the actual anomalies in the test set may differ significantly from the synthetic ones introduced through data augmentation. When such a mismatch occurs, the accuracy of NSEM's predictions can be compromised.
To mitigate this limitation, as many prior studies have shown that evolving data distributions often follow certain temporal patterns~\cite{edg2022,edg2023,edg2024}, we aim to learn such evolution pattern of the normal edge distribution as auxiliary information to guide the estimation process. By doing so, NSEM is no longer solely dependent on the data augmentation strategy, but can leverage historical trends in normality to improve robustness and generalization. To this end, we employ a Gated Recurrent Unit (GRU)~\cite{gru2014} to model the evolution pattern of normal edge distributions and update the hidden states $h_{s}^t$ in Equation~\eqref{NSEMmlp}: 
%%%%%%%%%%%%%%%%%%%%%%%%%%%%%%%%%%%%%%%%%%%%%%%%%%%%%%%%%%%%%%%%%%%%%%%%%%%%%%%%%
\begin{align}
    &\boldsymbol{x}_{s}^t=\boldsymbol{\mu}_{\text{norm}}^t \oplus 
    \text{Flatten}(\text{Chol}(\boldsymbol{\Sigma}_{\text{norm}}^t)),\label{gru1}\\
    &\boldsymbol{z}_{s}^{t} = \sigma (\boldsymbol{W}_{xz}\boldsymbol{x}_{s}^t + \boldsymbol{W}_{hz}\boldsymbol{h}_{s}^t),\label{gru2}\\
    &\boldsymbol{r}_{s}^{t} = \sigma (\boldsymbol{W}_{xr}\boldsymbol{x}_{s}^t + \boldsymbol{W}_{hr}\boldsymbol{h}_{s}^t),\label{gru3}\\
    &{\boldsymbol{h}^{'}}_{s}^{t} = \tanh(\boldsymbol{W}_{xh}\boldsymbol{x}_{s}^t + \boldsymbol{W}_{hh}\boldsymbol{h}_{s}^t),\label{gru4}\\
    &\boldsymbol{h}_{s}^t = (1-\boldsymbol{z}_{s}^t)\circ \boldsymbol{h}_{s}^{t-1}+\boldsymbol{z}_{s}^{t}\circ{\boldsymbol{h}^{'}}_{s}^{t},\label{gru5}
\end{align}
where $\circ$ denotes the Hadamard product and $\sigma(\cdot)$ represents the sigmoid function.

\vspace{1ex}\noindent\textbf{Whitening Transformation.} In the testing phase, we apply the predicted normal statistics to perform a whitening transformation~\cite{whitenning} on each edge embedding, aligning it to a standard Gaussian distribution. The whitening transformation is computed as 
\begin{equation}
    \widetilde{\boldsymbol{z}}_{ij}^t=({\hat{\boldsymbol{\Sigma}}_{\text{norm}}^{t}})^{-\frac{1}{2}}(\boldsymbol{z}_{ij}^t-\hat{\boldsymbol{\mu}}_{\text{norm}}^t),\label{whitening1}
\end{equation}
where $\boldsymbol{z}_{ij}^t$ denotes the edge embedding of edge $e_{ij}^t$ and $\widetilde{\boldsymbol{z}}_{ij}^t$ denotes the whitened edge embedding. $({\hat{\boldsymbol{\Sigma}}_{\text{norm}}^{t}})^{-\frac{1}{2}}$ denotes the inverse square root of the covariance matrix ${\hat{\boldsymbol{\Sigma}}_{\text{norm}}^{t}}$.

Finally, we train NSEM by minimizing the difference between the predicted and ground-truth statistics of normal edges. The Statistics Loss is defined as 
\begin{equation}
    \mathcal{L}_{\text{statistics}}=\left\lVert \hat{\boldsymbol{\mu}}_{\text{norm}}^t - \boldsymbol{\mu}_{\text{norm}}^t\right\rVert _{2}^{2} + \left\lVert \hat{\boldsymbol{\Sigma}}_{\text{norm}}^t - \boldsymbol{\Sigma}_{\text{norm}}^t\right\rVert _{2}^{2}.\label{statisticsloss}
\end{equation}

\subsection{Spatial-Temporal Encoder}
While representation learning on dynamic graphs is an important and challenging problem in its own right, it is not the main focus of this work. Therefore, we adopt a simplified adaptation of GC-LSTM~\cite{gc-lstm2022} to capture the temporal and spatial information critical to anomaly detection, which we find sufficient for learning expressive edge embeddings in our anomaly detection framework.
At each timestamp $t$, the input and output of the Spatial-Temporal Encoder are computed as 
\begin{align}
    &\boldsymbol{x}^t=\text{GNN}_{x}(\boldsymbol{X}^t),\\
    &\boldsymbol{f}^t=\sigma(\text{GNN}_{xf}(\boldsymbol{x}^t,\boldsymbol{A}^t)+ \text{GNN}_{hf}(\boldsymbol{h}^{t-1},\boldsymbol{A}^{t-1})),\label{glstm1}\\
    &\boldsymbol{i}^t=\sigma(\text{GNN}_{xi}(\boldsymbol{x}^t,\boldsymbol{A}^t)+\text{GNN}_{hi}(\boldsymbol{h}^{t-1},\boldsymbol{A}^{t-1})),\label{glstm2}\\
    &\boldsymbol{o}^t=\sigma(\text{GNN}_{xo}(\boldsymbol{x}^{t},\boldsymbol{A}^{t})+\text{GNN}_{ho}(\boldsymbol{h}^{t-1},\boldsymbol{A}^{t-1})),\label{glstm3}\\
    &\boldsymbol{z}^t=\tanh(\text{GNN}_{xz}(\boldsymbol{x}^{t},\boldsymbol{A}^{t})+\text{GNN}_{hz}(\boldsymbol{h}^{t-1},\boldsymbol{A}^{t-1})),\label{glstm4}\\
    &\boldsymbol{c}^t=(\boldsymbol{f}^t\boldsymbol{c}^{t-1}) + (\boldsymbol{i}^t\boldsymbol{z}^t), \quad \boldsymbol{n}^t=(\boldsymbol{f}^t\boldsymbol{n}^{t-1}) + \boldsymbol{i}^t,\label{glstm5}\\
    &\boldsymbol{h}^t=\boldsymbol{o}^t(\boldsymbol{c}^t/\boldsymbol{n}^t), \quad \boldsymbol{Z}_{\text{node}}^t = \boldsymbol{h}^t,\label{glstm6}
\end{align}
where $\text{GNN}(\cdot)$ denotes a graph neural network, $\boldsymbol{X}^{t}$ and $\boldsymbol{A}^{t}$ are the feature matrix and adjacency matrix of the snapshot $\mathcal{G}^{t}$, respectively, and $\boldsymbol{Z}_{\text{node}}^{t} \in \mathbb{R}^{N^t \times d}$ represents the node embeddings at timestamp $t$. In addition, $\boldsymbol{n}^t$ is a normalizer state~\cite{xlstm2024} used to stabilize the hidden state $\boldsymbol{h}^t$, preventing large numerical fluctuations that could hinder NSEM from accurately learning and predicting the distributional statistics.
Let $\boldsymbol{z}_{i}^{t}$ denote the $i$-th row of $\boldsymbol{Z}_{\text{node}}^{t}$, representing the representation of node $i$ at timestamp $t$.
For an edge $e_{ij}^{t} \in \mathcal{E}^{t}$, its edge embedding $\boldsymbol{z}_{ij}^{t}$ is computed as $\boldsymbol{z}_{ij}^t=\boldsymbol{z}_{i}^t+\boldsymbol{z}_{j}^t$.

The Spatial-Temporal Encoder is trained via a graph reconstruction task, where we aim to reconstruct both the feature matrix $\boldsymbol{X}^{t}$ and the adjacency matrix $\boldsymbol{A}^{t}$ from the learned node embeddings $\boldsymbol{Z}_{\text{node}}^{t}$. The reconstructions are given by:
\begin{align}
    \hat{\boldsymbol{X}}^t=\text{MLP}(\boldsymbol{Z}_{\text{node}}^t), \quad \hat{\boldsymbol{A}}^t=\sigma(\boldsymbol{Z}_{\text{node}}^t{\boldsymbol{Z}_{\text{node}}^t}^{\top}).\label{recon}
\end{align}
Then the reconstruction loss is defined as
\begin{equation}
    \mathcal{L}_{\text{recon}}=\left\lVert\hat{\boldsymbol{X}}^t-\boldsymbol{X}^t \right\rVert_{F}^{2} + \left\lVert\hat{\boldsymbol{A}}^t-\boldsymbol{A}^t \right\rVert_{F}^{2},\label{reconloss}
\end{equation}
where $\left\lVert\cdot\right\rVert_{F}$ denotes the Frobenius norm.

\subsection{Anomaly Detector}
After obtaining the node embeddings from the Spatial-Temporal Encoder and computing the edge embeddings, we utilize the predicted normal edge statistics from NSEM to perform whitening on the edge embeddings using Equation~\eqref{whitening1}.
The whitened edge embeddings are then fed into the Anomaly Detector to compute anomaly scores:
\begin{equation}
    f(e_{ij}^t)=\sigma(\phi_{a}(\tilde{\boldsymbol{z}}_{ij}^t)+\boldsymbol{b}),\label{anoscore}
\end{equation}
where $\phi_{a}(\cdot)$ is an MLP, $\boldsymbol{b}$ is a learnable bias term, and $\sigma(\cdot)$ denotes the sigmoid activation function.

Following previous works~\cite{taddy2021,rustgraph2024}, we employ negative sampling to generate pseudo-anomalous edges for training.
Specifically, pseudo anomalies are created by either randomly connecting nonexistent edges~\cite{taddy2021} or randomly altering the destination nodes of existing edges~\cite{addgraph2019}. With the generated pseudo labels, the Anomaly Detector is trained using a binary cross-entropy (BCE) loss defined as
\begin{equation}
    \mathcal{L}_{\text{bce}}=-\sum_{e_{ij}^t\in\mathcal{E}^t}\{(1-y_{ij}^t)\log(1-f(e_{ij}^t))+y_{ij}^t\log(f(e_{ij}^t))\},\label{lossbce}
\end{equation}
where $y_{ij}^t$ is the given pseudo label of the edge $e_{ij}^t$.

\subsection{Training Procedure}
To avoid the instability caused by continuously changing embeddings and their corresponding statistics during training, we do not train WhENDS in an end-to-end manner. Instead, we adopt a sequential training strategy, where each module is trained independently.

We begin by training the Spatial-Temporal Encoder on the original training set by minimizing the reconstruction loss $\mathcal{L}_{\text{recon}}$, until it can effectively capture the spatial-temporal structure of dynamic graphs.
Once the encoder is trained, we apply the data augmentation strategy described earlier:
For each snapshot in the training set, we randomly insert 1\% to 10\% of synthetic edges to simulate the anomalous edge distribution $\mathcal{D}_{\text{ano}}^t$ encountered during testing. During the training of NSEM, for each timestamp $t$, we compute the statistics of overall edge embeddings $(\mu_{\text{all}}^t, \Sigma_{\text{all}}^t)$ and the ground-truth statistics of normal edge embeddings $(\mu_{\text{norm}}^t, \Sigma_{\text{norm}}^t)$ by
\begin{align}
        &\boldsymbol{\mu}_{\text{all}}^t=\frac{1}{|\mathcal{E}^t|}\sum_{e_{ij}^t\in\mathcal{E}^t}\boldsymbol{z}_{ij}^t,\label{allmu}\\
        &\boldsymbol{\Sigma}_{\text{all}}^t=\frac{1}{|\mathcal{E}^t|}\sum_{e_{ij}^t\in\mathcal{E}^t}(\boldsymbol{z}_{ij}^t-\boldsymbol{\mu}_{\text{all}}^t)(\boldsymbol{z}_{ij}^t-\boldsymbol{\mu}_{\text{all}}^t)^{\top},\label{allsigma}\\
        &\boldsymbol{\mu}_{\text{norm}}^t=\frac{1}{|\mathcal{E}_{\text{norm}}^t|}\sum_{e_{ij}^t\in\mathcal{E}_{\text{norm}}^t}\boldsymbol{z}_{ij}^t,\label{normmu}\\
        &\boldsymbol{\Sigma}_{\text{norm}}^t=\frac{1}{|\mathcal{E}_{\text{norm}}^t|}\sum_{e_{ij}^t\in\mathcal{E}_{\text{norm}}^t}(\boldsymbol{z}_{ij}^t-\boldsymbol{\mu}_{\text{norm}}^t)(\boldsymbol{z}_{ij}^t-\boldsymbol{\mu}_{\text{norm}}^t)^{\top}.\label{normsigma}
\end{align}
The $(\boldsymbol{\mu}_{\text{all}}^t, \boldsymbol{\Sigma}_{\text{all}}^t)$ are then used as inputs to NSEM to predict $(\hat{\boldsymbol{\mu}}_{\text{norm}}^t,\hat{\boldsymbol{\Sigma}}_{\text{norm}}^t)$, and $(\boldsymbol{\mu}_{\text{norm}}^t, \boldsymbol{\Sigma}_{\text{norm}}^t)$ are used to calculate $\mathcal{L}_{\text{statistics}}$ by Equation~\eqref{statisticsloss} to train NSEM.

Finally, for Anomaly Detector training, we generate pseudo-anomalous edges via negative sampling on the original training set.
Unlike in the testing phase, where the whitening transformation is performed using the predicted statistics from NSEM, during training, we directly use the ground-truth statistics calculated by Equations~\eqref{normmu}--\eqref{normsigma} to whiten edge embeddings:
\begin{align}
\widetilde{\boldsymbol{z}}_{ij}^t=({\boldsymbol{\Sigma}_{\text{norm}}^{t}})^{-\frac{1}{2}}(\boldsymbol{z}_{ij}^t-\boldsymbol{\mu}_{\text{norm}}^t).\label{whitening2}
\end{align}
The whitened edge embeddings are then fed into the Anomaly Detector to compute anomaly scores by Equation~\eqref{anoscore}, and the model is trained using the binary cross-entropy loss $\mathcal{L}_{\text{bce}}$ with pseudo labels, which are calculated by Equation~\eqref{lossbce}.

Since $\mathcal{L}_{\text{recon}}$, $\mathcal{L}_{\text{statistics}}$, and $\mathcal{L}_{\text{bce}}$ are used to independently train different components, no loss weight tuning is required. We also summarize the overall training procedure in Algorithm~\ref{algorithm}.

\begin{algorithm}[t]
\caption{Training Procedure of WhENDS}
\label{algorithm}
\begin{algorithmic}[1]
\Require Dynamic Graph $\mathbb{G} = \{\mathcal{G}^{1},\dots,\mathcal{G}^{T}\}$, training epochs $I_{e}, I_{n}, I_{a}$
\State Initialize network parameters;
\For {$i=1$ to $I_{e}$}
    \For {$t=1$ to $T$}
        \State Encode graph representation using Equations~\eqref{glstm1}--\eqref{glstm6}
        \State Calculate $\mathcal{L}_{\text{recon}}$ using Equations~\eqref{recon}--\eqref{reconloss}
        \State Backpropagate and update parameters of the Spatial-Temporal Encoder
    \EndFor
\EndFor
\State Inject edges into $\mathbb{G}$
\For {$i=1$ to $I_{n}$}
    \For {$t=1$ to $T$}
        \State Encode graph representation using Equations~\eqref{glstm1}--\eqref{glstm6}
        \State Calculate statistics of overall edge embeddings using Equations~\eqref{allmu}--\eqref{allsigma}
        \State Calculate statistics of normal edge embeddings using Equations~\eqref{normmu}--\eqref{normsigma}
        \State Predict normal statistics using Equations~\eqref{NSEMmlp}--\eqref{predsigma}
        \State Update hidden state using Equations~\eqref{gru1}--\eqref{gru5}
        \State Calculate $\mathcal{L}_{\text{statistics}}$ using Equation~\eqref{statisticsloss}
        \State Backpropagate and update parameters of NSEM
    \EndFor
\EndFor
\State Generate pseudo anomalies in $\mathbb{G}$
\For {$i=1$ to $I_{a}$}
    \For {$t=1$ to $T$}
        \State Encode graph representation using Equations~\eqref{glstm1}--\eqref{glstm6}
        \State Calculate statistics of normal edge embeddings using Equations~\eqref{normmu}--\eqref{normsigma}
        \State Whiten edge embeddings using Equation~\eqref{whitening2}
        \State Calculate anomaly score $f(e_{ij}^t)$ using Equation~\eqref{anoscore}
        \State Calculate $\mathcal{L}_{\text{bce}}$ using Equation~\eqref{lossbce}
        \State Backpropagate and update parameters of Anomaly Detector
    \EndFor
\EndFor
\end{algorithmic}
\end{algorithm}

\subsection{Complexity Analysis}
In this subsection, we briefly analyze the time complexity of each module of WhENDS. 

For the Spatial-Temporal Encoder, the time complexity for an $L$-layer GraphSAGE~\cite{graphsage2018}, which samples uniformly a fixed number $k$ of 1-hop neighbors, is $O(Ld^2k^L)$~\cite{rustgraph2024}. In a single-step LSTM update, we perform computations using GraphSAGE and matrix multiplications. The time complexity for one update is therefore $O(Ld^2k^L + d^2) = O(Ld^2k^L)$. 

For reconstruction, given a graph with $n$ nodes, the complexity is $O(nd^2 + n^2d)$ per snapshot, resulting in a total reconstruction cost of $O((nd^2 + n^2d)T)$ over $T$ timestamps.

For the NSEM, we flatten the covariance matrix and concatenate it with the mean vector, forming an input of dimension approximately $d^2$. The computational complexity of both the MLP and GRU operations is $O(d^4)$. For a snapshot with $e$ edges, performing whitening on all edge embeddings requires $O(ed^2)$. Thus, the total time complexity of NSEM is $O((d^2 + e)d^2T)$.

For the Anomaly Detector, given $T$ snapshots each with $e$ edges, the total complexity is $O(ed^2T)$.

Putting all components together, the overall time complexity of WhENDS is $O((Lk^L + e + n + d^2)d^2T)$. Since $Lk^L$ is typically small in practice, it can be considered negligible. Therefore, the simplified overall complexity of WhENDS becomes $O(ed^2T + nd^2T + d^4T)$.

\begin{table}[t]
\centering
\caption{Dataset statistics.}
\label{datasets}
\begin{tabular}{ccccc}
\toprule
\textbf{Dataset} & \textbf{Nodes} & \textbf{Edges} &\textbf{Avg.Degree} \\
\midrule
    UCI Messages & 1,899 & 13,838 & 14.57 \\
    Digg & 30,360 & 85,155 & 5.61 \\
    Bitcoin-Alpha & 3,777 & 24,173 & 12.80 \\
    Bitcoin-OTC & 5,881 & 35,588 & 12.10 \\
\bottomrule
\end{tabular}
\end{table}

\begin{table*}[t]
    \renewcommand{\arraystretch}{1.2}
    \centering
\caption{Anomaly detection results reported using the AUC-ROC score. The best result in each column is highlighted in bold, and the second-best is underlined.}
    \label{maintable}
    \begin{tabular}{ccccccccccccc}
    \toprule
    \multirow{2}{*}{Methods} & \multicolumn{3}{c}{UCI Messages} & \multicolumn{3}{c}{Digg} & \multicolumn{3}{c}{Bitcoin-Alpha} & \multicolumn{3}{c}{Bitcoin-OTC} \\
    \cmidrule{2-4} \cmidrule{5-7} \cmidrule{8-10} \cmidrule{11-13}
    ~ &1\% &5\% &10\% &1\% &5\% &10\%&1\% &5\% &10\% &1\% &5\% &10\%\\
    \midrule
    Sedanspot & 0.7342 & 0.7156 & 0.7061 & 0.6976 & 0.6784 & 0.6396 & 0.7380 & 0.7264 & 0.7085 & 0.7346 & 0.7284 & 0.7156\\
    CM-Sketch & 0.7320 & 0.6968 & 0.6835 & 0.6884 & 0.6675 & 0.6358 & 0.7146 & 0.7015 & 0.6887 & 0.7412 & 0.7338 & 0.7242\\
    \midrule
    Spectral Clustering & 0.6324 & 0.6104 & 0.5794 & 0.5949 & 0.5823 & 0.5591 & 0.7401 & 0.7275 & 0.7167 & 0.7624 & 0.7376 & 0.7047\\
    DeepWalk & 0.7514 & 0.7391 & 0.6979 & 0.7080 & 0.6881 & 0.6396 & 0.6985 & 0.6874 & 0.6793 & 0.7423 & 0.7356 & 0.7287\\
    \midrule
    NetWalk & 0.7758 & 0.7647 & 0.7226 & 0.7563 & 0.7176 & 0.6837 & 0.8385 & 0.8357 & 0.8350 & 0.7785 & 0.7694 & 0.7534\\
    AddGraph & 0.8083 & 0.8090 & 0.7688 & 0.8341 & 0.8470 & 0.8369 & 0.8665 & 0.8403 & 0.8498 & 0.8352 & 0.8455 & 0.8592\\
    StrGNN & 0.8179 & 0.8252 & 0.7959 & 0.8162 & 0.8254 & 0.8272 & 0.8574 & 0.8667 & 0.8627 & 0.9012 & 0.8775 & 0.8836\\
    TADDY & 0.8912 & 0.8398 & 0.8370 & 0.8617 & 0.8545 & 0.8440 & \underline{0.9451} & 0.9341 & \textbf{0.9423} & 0.9455 & 0.9340 & 0.9425\\
    RustGraph & \underline{0.9128} & \underline{0.9117} & \underline{0.9124} & \underline{0.8795} & \underline{0.8577} & \underline{0.8624} & 0.9447 & \underline{0.9348} & 0.9207 & \underline{0.9608} & \underline{0.9578} & \underline{0.9603}\\
    \midrule
    WhENDS & \textbf{0.9660} & \textbf{0.9627} & \textbf{0.9584} & \textbf{0.8810} & \textbf{0.8708} & \textbf{0.8678} & \textbf{0.9736} & \textbf{0.9610} & \underline{0.9416} & \textbf{0.9721} & \textbf{0.9728} & \textbf{0.9713}\\
    \bottomrule
    \end{tabular}
\end{table*}

\section{Experiments}
\subsection{Experiment Setup}
% \subsubsection{Datasets}
% \ 
% \newline

\noindent \textbf{Datasets.} We evaluate the performance of our model on four widely used dynamic graph anomaly detection datasets. Summary statistics are provided in Table~\ref{datasets}.

\textit{UCI Messages}~\cite{uci} records a directed communication network among users of an online student community at the University of California, Irvine. Each node represents a user, and each directed edge corresponds to a message sent from one user to another.

\textit{Digg}~\cite{digg} captures user interactions on the social news platform Digg. Nodes correspond to users, and a directed edge indicates that one user replied to another user’s comment or post.

\textit{Bitcoin-Alpha}~\cite{btcalpha} represents a who-trusts-whom network collected from the Bitcoin-Alpha trading platform. Each node denotes a trader, and an edge indicates that a user assigned a trust rating to another user.

\textit{Bitcoin-OTC}~\cite{btcotc} is another who-trusts-whom network, collected from the Bitcoin OTC trading platform. Similar to Bitcoin-Alpha, nodes represent users and edges indicate directed trust ratings between users.

% \subsubsection{Baselines}
% \ 
% \newline
\vspace{1ex}
\noindent \textbf{Baselines.} We compare our method against nine state-of-the-art baseline approaches, which can be grouped into the following three categories:

(1) \textit{Classical Anomaly Scoring Methods:} \textbf{Sedanspot}~\cite{sedanspot} detects anomalies using a random walk-based holistic scoring function that captures changes in connectivity patterns. \textbf{CM-Sketch}~\cite{cmsketch2016} applies the Count-Min Sketch algorithm to approximate empirical anomaly metrics and identify anomalous edges.

(2) \textit{Graph Embedding Methods:} \textbf{Spectral Clustering}~\cite{spectralclustering} learns node embeddings by maximizing local neighborhood similarity via spectral decomposition of the adjacency matrix. \textbf{DeepWalk}~\cite{deepwalk2014} generates node embeddings using the Skip-gram model applied to random walks sampled from the graph structure.

(3) \textit{Deep Anomaly Detection Methods:} \textbf{NetWalk}~\cite{netwalk2019} adopts a reconstruction-based strategy, training an autoencoder on random walk sequences to generate node and edge embeddings, and computing anomaly scores via clustering. \textbf{AddGraph}~\cite{addgraph2019} constructs $h$-hop enclosing subgraphs for each edge and applies a spatial-temporal model to learn discriminative edge embeddings. \textbf{TADDY}~\cite{taddy2021} employs edge-based substructure sampling combined with advanced node embeddings as input to a transformer-based model for dynamic anomaly detection. \textbf{RustGraph}~\cite{rustgraph2024} integrates GNNs with GRUs to capture spatial-temporal dependencies and leverages graph contrastive learning to encode dynamic graphs.

% \subsubsection{Experiment Details}
% \ 
% \newline
\vspace{1ex}
\noindent\textbf{Experiment Details.} We follow previous works~\cite{rustgraph2024} for data preprocessing and use node2vec~\cite{node2vec} to generate raw features for each node.
For each dataset, the first 50\% of the snapshots are used as the training set, while the remaining 50\% are used as the testing set. To evaluate model performance under different anomaly rates, we inject anomalous edges into the test set with three different proportions (1\%, 5\%, 10\%).

% \subsubsection{Metric}
% \ 
% \newline
\vspace{1ex}
\noindent\textbf{Metric.} We adopt the Area Under the Receiver Operating Characteristic Curve (AUC-ROC) as the primary metric to evaluate model performance.
The AUC-ROC score ranges from 0 to 1, where a higher score indicates better anomaly detection performance.

% \subsubsection{Parameter Setting}
% \ 
% \newline
\vspace{1ex}
\noindent\textbf{Parameter Setting.} For all datasets, we set the embedding dimension to $d = 128$.
To ensure fair comparisons, following previous work~\cite{netwalk2019}, we set the snapshot size to $1{,}000$ for UCI Messages and Bitcoin-OTC, $2{,}000$ for Bitcoin-Alpha, and $6{,}000$ for Digg. 

All GNN and MLP components are implemented with two layers, and we use GraphSAGE~\cite{graphsage2018} as the GNN encoder in our work. The NSEM utilizes a two-layer GRU, and the Spatial-Temporal Encoder is constructed with two layers of the LSTM-based structure described earlier.

During training, the hyperparameters are configured as follows:

- \textbf{Spatial-Temporal Encoder:}  
  UCI Messages and Digg: 200 epochs, learning rate $0.0005$;  
  Bitcoin-Alpha: 100 epochs, learning rate $0.0005$;  
  Bitcoin-OTC: 100 epochs, learning rate $0.0001$.

- \textbf{NSEM:}  
  All datasets: 200 epochs, learning rate $0.001$.

- \textbf{Anomaly Detector:}  
  UCI Messages: 800 epochs, learning rate $0.001$;  
  Digg: 1600 epochs, learning rate $0.001$;  
  Bitcoin-Alpha: 800 epochs, learning rate $0.001$;  
  Bitcoin-OTC: 1600 epochs, learning rate $0.0005$.

All modules are optimized using the Adam optimizer, with the weight decay rate set to $0.9$.

% \subsubsection{Computing Infrastructures}
% \ 
% \newline
\vspace{1ex}
\subsection{Anomaly Detection Results}
Anomaly detection results on the four datasets are reported in Table~\ref{maintable}, from which we draw the following observations:
\begin{itemize}
    \item WhENDS achieves state-of-the-art results in most of the 12 experimental settings. Compared with the best-performing baselines, it yields an average improvement of $2.36\%$, demonstrating the effectiveness of our method and underscoring the importance of addressing NDS in anomaly detection.
    
    \item The performance gain of WhENDS varies across datasets. The average improvements on UCI Messages, Digg, Bitcoin-Alpha, and Bitcoin-OTC are $5.38\%$, $0.77\%$, $1.91\%$, and $1.29\%$, respectively. This variation is likely due to different degrees of NDS present in each dataset.
    
    \item WhENDS shows larger improvements in settings with sparser anomalies. At anomaly ratios of $1\%$, $5\%$, and $10\%$, the average AUC-ROC gains are $2.56\%$, $2.88\%$, and $1.65\%$, respectively. We attribute this to WhENDS’s whitening-based alignment, which effectively standardizes normal edge distributions across time, enhancing its ability to detect subtle anomalies even when they are scarce.
\end{itemize}

\begin{table*}[t]
    \renewcommand{\arraystretch}{1.2}
    \centering
    \caption{The ablation study results are reported using the AUC-ROC score. The best result in each column is highlighted in bold, and the second-best is underlined.}
    \label{ablationstudy}
    \begin{tabular}{ccccccccccccc}
    \toprule
    \multirow{2}{*}{Methods} & \multicolumn{3}{c}{UCI Messages} & \multicolumn{3}{c}{Digg} & \multicolumn{3}{c}{Bitcoin-Alpha} & \multicolumn{3}{c}{Bitcoin-OTC} \\
    \cmidrule{2-4} \cmidrule{5-7}\cmidrule{8-10} \cmidrule{11-13}
    ~ &1\% &5\% &10\% &1\% &5\% &10\% &1\% &5\% &10\% &1\% &5\% &10\%\\
    \midrule
    WhENDS & \underline{0.9660} & \textbf{0.9627} & \textbf{0.9584} & \underline{0.8810} & \textbf{0.8708} & \underline{0.8678} & \textbf{0.9736} & \textbf{0.9610} & \textbf{0.9416} & \underline{0.9721} & \textbf{0.9728} & \textbf{0.9713} \\
    w/o ST Encoder & 0.9315 & 0.9317 & 0.9264 & 0.8533 & 0.8514 & 0.8562 & 0.9486 & 0.9360 & 0.9160 & 0.9430 & 0.9483 & 0.9369 \\
    w/o GRU & \textbf{0.9669} & \underline{0.9626} & \underline{0.9582} & \textbf{0.8811} & \underline{0.8705} & \textbf{0.8684}& \underline{0.9633} & \underline{0.9468} & \underline{0.9388}& \textbf{0.9730} & \underline{0.9727} & \underline{0.9711}\\
    w/o DataAug & 0.9315 & 0.9317 & 0.9264 & 0.8651 & 0.8557 & 0.8566 & 0.9450 & 0.9262 & 0.9133 &0.9538 & 0.9456 & 0.9567 \\
    w/o NSEM & 0.9170 & 0.9137 & 0.9132 & 0.8514 & 0.8319 & 0.8386 & 0.9407 & 0.9287 & 0.9125 & 0.9345 & 0.9375 & 0.9367 \\
    \bottomrule
    \end{tabular}
\end{table*}

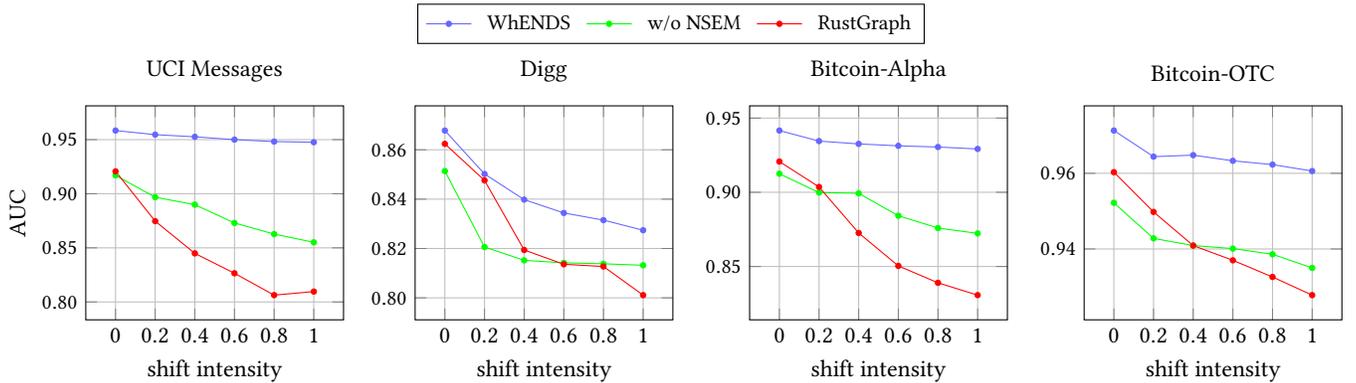
\begin{figure*}
    \begin{center}
        \ref{sharedlegend}
    \end{center}
    \centering
    % 第一行
    \begin{subfigure}{0.23\textwidth}
        \centering
        \begin{tikzpicture}
            \begin{axis}[
                scale=0.5,
                xtick={0,0.2,0.4,0.6,0.8,1.0},
                yticklabel style={
                    font=\small,
                    /pgf/number format/fixed,
                    /pgf/number format/precision=2,
                    /pgf/number format/fixed zerofill
                },
                xlabel near ticks,
                ylabel near ticks,
                title={UCI Messages},
                xlabel={shift intensity},
                ylabel={AUC},
                grid=major,
                % legend pos=south west,
                enlargelimits=0.15,
                legend to name=sharedlegend,
                legend style={
                    legend columns=-1,
                    column sep=1ex,
                    font=\small,
                    cells={anchor=west},
                    at={(0.5,1.25)},
                    anchor=south,
                }
            ]
            \addplot[
                color=blue!60,
                mark=*,
                mark size=1pt
                ]
                coordinates {
                (0, 0.9584) (0.2,0.9546) (0.4,0.9526) (0.6,0.9500) (0.8,0.9482) (1.0,0.9476)
                };
            % \addlegendentry{Ours}
        
            \addplot[
                color=green,
                mark=*,
                mark size=1pt
                ]
                coordinates {
                (0,0.9170) (0.2,0.8968) (0.4,0.8899) (0.6,0.8730) (0.8,0.8627) (1.0,0.8551)
                };
            % \addlegendentry{w/o WhENDS}
        
            \addplot[
                color=red,
                mark=*,
                mark size=1pt
                ]
                coordinates {
                (0,0.9207) (0.2,0.8747) (0.4,0.8449) (0.6,0.8265) (0.8,0.8063) (1.0,0.8096)
                };
            % \addlegendentry{RustGraph}
            \addlegendimage{blue, mark=*}
            \addlegendentry{WhENDS}
            \addlegendimage{green, mark=*}
            \addlegendentry{w/o NSEM}
            \addlegendimage{red, mark=*}
            \addlegendentry{RustGraph}
            \end{axis}
        \end{tikzpicture}
        % \caption{子图 1}
    \end{subfigure}
    \hspace{8pt}
    \hfill
    \begin{subfigure}{0.23\textwidth}
        \centering
        \begin{tikzpicture}
            \begin{axis}[
                scale=0.5,
                xtick={0,0.2,0.4,0.6,0.8,1.0},
                yticklabel style={
                    font=\small,
                    /pgf/number format/fixed,
                    /pgf/number format/precision=2,
                    /pgf/number format/fixed zerofill
                },
                xlabel near ticks,
                % ylabel near ticks,
                title={Digg},
                xlabel={shift intensity},
                grid=major,
                % legend pos=south west,
                enlargelimits=0.15,
                legend style={
                    at={(0.5,-0.4)},
                    anchor=north,
                    font=\small,
                }
            ]
            \addplot[
                color=blue!60,
                mark=*,
                mark size=1pt
                ]
                coordinates {
                (0,0.8678) (0.2,0.8502) (0.4,0.8398) (0.6,0.8344) (0.8,0.8315) (1.0,0.8274)
                };
            % \addlegendentry{Ours}
        
            \addplot[
                color=green,
                mark=*,
                mark size=1pt
                ]
                coordinates {
                (0,0.8514) (0.2,0.8206) (0.4,0.8152) (0.6,0.8142) (0.8,0.8138) (1.0,0.8132)
                };
            % \addlegendentry{w/o WhENDS}
        
            \addplot[
                color=red,
                mark=*,
                mark size=1pt
                ]
                coordinates {
                (0,0.8624) (0.2,0.8476) (0.4,0.8194) (0.6,0.8136) (0.8,0.8127) (1.0,0.8011)
                };
            \end{axis}
        \end{tikzpicture}
        % \caption{子图 2}
    \end{subfigure}
    \hfill
    \begin{subfigure}{0.23\textwidth}
        \centering
        \begin{tikzpicture}
            \begin{axis}[
                scale=0.5,
                xtick={0,0.2,0.4,0.6,0.8,1.0},
                yticklabel style={
                    font=\small,
                    /pgf/number format/fixed,
                    /pgf/number format/precision=2,
                    /pgf/number format/fixed zerofill
                },
                xlabel near ticks,
                % ylabel near ticks,
                title={Bitcoin-Alpha},
                xlabel={shift intensity},
                grid=major,
                % legend pos=south west,
                enlargelimits=0.15,
                legend style={
                    at={(0.5,-0.4)},
                    anchor=north,
                    font=\small,
                }
            ]
            \addplot[
                color=blue!60,
                mark=*,
                mark size=1pt
                ]
                coordinates {
                (0,0.9416) (0.2,0.9345) (0.4,0.9326) (0.6,0.9313) (0.8,0.9305) (1.0,0.9292)
                };
            % \addlegendentry{Ours}
        
            \addplot[
                color=green,
                mark=*,
                mark size=1pt
                ]
                coordinates {
                (0,0.9125) (0.2,0.8998) (0.4,0.8993) (0.6,0.8842) (0.8,0.8759) (1.0,0.8723)
                };
            % \addlegendentry{w/o WhENDS}
        
            \addplot[
                color=red,
                mark=*,
                mark size=1pt
                ]
                coordinates {
                (0,0.9207) (0.2,0.9036) (0.4,0.8726) (0.6,0.8504) (0.8,0.8390) (1.0,0.8307)
                };
            % \addlegendentry{RustGraph}
            % \addlegendimage{blue, mark=*}
            % \addlegendentry{Ours}
            % \addlegendimage{green, mark=*}
            % \addlegendentry{w/o NSEM}
            % \addlegendimage{red, mark=*}
            % \addlegendentry{RustGraph}
            \end{axis}
        \end{tikzpicture}
        % \caption{子图 3}
    \end{subfigure}
    \hfill
    \begin{subfigure}{0.23\textwidth}
        \centering
        \begin{tikzpicture}
            \begin{axis}[
                scale=0.5,
                xtick={0,0.2,0.4,0.6,0.8,1.0},
                yticklabel style={
                    font=\small,
                    /pgf/number format/fixed,
                    /pgf/number format/precision=2,
                    /pgf/number format/fixed zerofill
                },
                xlabel near ticks,
                % ylabel near ticks,
                title={Bitcoin-OTC},
                xlabel={shift intensity},
                grid=major,
                % legend pos=south west,
                enlargelimits=0.15,
                legend style={
                    at={(0.5,-0.4)},
                    anchor=north,
                    font=\small,
                }
            ]
            \addplot[
                color=blue!60,
                mark=*,
                mark size=1pt
                ]
                coordinates {
                (0,0.9713) (0.2,0.9644) (0.4,0.9648) (0.6,0.9633) (0.8,0.9623) (1.0,0.9606)
                };
            % \addlegendentry{Ours}
        
            \addplot[
                color=green,
                mark=*,
                mark size=1pt
                ]
                coordinates {
                (0,0.9522) (0.2,0.9428) (0.4,0.9409) (0.6,0.9401) (0.8,0.9386) (1.0,0.9350)
                };
            % \addlegendentry{w/o WhENDS}
        
            \addplot[
                color=red,
                mark=*,
                mark size=1pt
                ]
                coordinates {
                (0, 0.9603) (0.2,0.9498) (0.4,0.9409) (0.6,0.9370) (0.8,0.9326) (1.0,0.9278)
                };
            % \addlegendentry{RustGraph}
            % \addlegendimage{blue, mark=*}
            % \addlegendentry{Ours}
            % \addlegendimage{green, mark=*}
            % \addlegendentry{w/o NSEM}
            % \addlegendimage{red, mark=*}
            % \addlegendentry{RustGraph}
            \end{axis}
        \end{tikzpicture}
        % \caption{子图 4}
    \end{subfigure}
    % \vspace{10em}

    \caption{The AUC-ROC curves under differnent intensity of normality distribution shift.}
    \label{ndsexp}
\end{figure*}

\subsection{Ablation Study}
We conduct an ablation study to investigate the contribution of each component in our model. 
Specifically, in \textbf{w/o ST Encoder}, we replace the Spatial-Temporal Encoder with a simple two-layer GNN, so the model can only capture structural information and loses the ability to model temporal dynamics. In \textbf{w/o GRU}, we remove the GRU component from NSEM, preventing it from learning the temporal evolution patterns of normal distributions. In \textbf{w/o DataAug}, we disable data augmentation during the training of NSEM. In this setting, NSEM directly predicts the normal statistics from the previous hidden state $h_s^{t-1}$, rather than predicting the deviation between the overall and normal edge embedding statistics. In \textbf{w/o NSEM}, we remove the entire NSEM module, and thus no whitening is applied to the edge embeddings, leaving the model unable to handle NDS.

The results of the ablation study are shown in Table~\ref{ablationstudy}, from which we draw the following conclusions:
\begin{itemize}
    \item NSEM is the most critical component of WhENDS. In all datasets, removing NSEM results in the worst performance among all ablated variants, leading to an average drop in AUC-ROC score of 3.69\%, which demonstrates the effectiveness of NSEM.

    \item Among the components of NSEM, data augmentation proves to be the most influential. Disabling data augmentation results in an average drop in AUC-ROC of 2.69\%, whereas removing the GRU while retaining data augmentation leads to negligible performance differences in UCI Messages, Digg, and Bitcoin-OTC. This suggests that, in most cases, NSEM is able to rely solely on data augmentation to make accurate predictions.

    % \item In Bitcoin-Alpha, removing the GRU results in a decrease in AUC-ROC of up to 1.42\%, indicating that modeling the evolution of the normal distribution is essential for accurate estimation in this dataset. Although using the GRU alone without data augmentation leads to a performance drop of 2.43\%, it still achieves competitive results, demonstrating that learning distributional evolution alone can provide reasonably effective predictions.
    \item In Bitcoin-Alpha, removing the GRU results in a decrease in AUC-ROC of up to 1.42\%. We attribute this to the mismatch between the synthetic anomalies introduced by data augmentation and the actual anomalous edges in the test set, as previously discussed. This result also suggests that learning evolution patterns of distribution helps mitigate the adverse effects caused by this mismatch. Moreover, although using the GRU alone without data augmentation leads to a performance drop of 2.43\%, it still achieves competitive results, demonstrating that learning distributional evolution alone can provide reasonably effective predictions.

    \item Replacing the Spatial-Temporal Encoder with a simple GNN causes an average decrease in AUC-ROC of 2.68\%, indicating the importance of capturing temporal information for anomaly detection in dynamic graphs, as also supported by previous studies~\cite{taddy2021,rustgraph2024}.
\end{itemize}

% NSEM is the most critical component of WhENDS.
% Removing NSEM leads to an average drop in AUC-ROC score of 3.69\%, highlighting the importance of addressing NDS in anomaly detection.
% More specifically, the decreases in UCI Messages, Digg, Bitcoin-Alpha, and Bitcoin-OTC are 4.77\%, 3.25\%, 3.14\%, and 3.58\%, respectively. This suggests that NDS is prevalent across different dynamic graph datasets, and the varying degrees of performance degradation may indicate the inherent strength of distribution shift in each dataset.

% Among the components of NSEM, data augmentation proves to be the most influential.
% Disabling data augmentation results in an average drop in AUC-ROC of 2.69\%, whereas removing the GRU while retaining data augmentation leads to negligible performance difference in UCI Messages, Digg and Bitcoin-OTC.

% In Bitcoin-Alpha, removing the GRU results in a decrease in AUC-ROC of up to 1.42\%, indicating that modeling the evolution of the normal distribution is essential for accurate estimation in this dataset. Although using GRU alone without data augmentation leads to a performance drop of 2.43\%, it still achieves competitive results, demonstrating that learning distributional evolution alone can provide reasonably effective predictions.

% Replacing the Spatial-Temporal Encoder with a simple GNN causes an average decrease in AUC-ROC of 2.68\%, indicating the importance of capturing temporal information for anomaly detection in dynamic graphs, as also supported by previous studies~\cite{taddy2021,rustgraph2024}.
\begin{figure*}
    \centering
    % 第一行
    \begin{subfigure}{0.23\textwidth}
        \centering
        \begin{tikzpicture}
            \begin{axis}[
                scale=0.5,
                xmode=log,             
                log basis x=2,
                xtick={2,4,8,16,32,64,128},
                xticklabels={2,4,8,16,32,64,128},
                yticklabel style={
                    font=\small,
                    /pgf/number format/fixed,
                    /pgf/number format/precision=2,
                    /pgf/number format/fixed zerofill
                },
                xlabel near ticks,
                ylabel near ticks,
                title={UCI Messages},
                xlabel={embedding dimension},
                ylabel={AUC},
                grid=major,
                legend pos=south east,
                enlargelimits=0.15,
                legend style={font=\small, fill=none, inner sep=1pt}
            ]
            \addplot[
                color=blue!60,
                mark=*,
                mark size=1pt
                ]
                coordinates {
                (2, 0.7844) (4,0.8673) (8,0.8905) (16,0.8982) (32,0.9324) (64,0.9453) (128,0.9660)
                };
            \addlegendentry{1\%}
        
            \addplot[
                color=green,
                mark=*,
                mark size=1pt
                ]
                coordinates {
                (2, 0.7909) (4,0.8815) (8,0.8845) (16,0.9131) (32,0.9220) (64,0.9415) (128,0.9630)
                };
            \addlegendentry{5\%}
        
            \addplot[
                color=red,
                mark=*,
                mark size=1pt
                ]
                coordinates {
                (2, 0.7867) (4,0.8778) (8,0.8872) (16,0.9130) (32,0.9187) (64,0.9434) (128,0.9582)
                };
            \addlegendentry{10\%}
            \end{axis}
        \end{tikzpicture}
        % \caption{子图 1}
    \end{subfigure}
    \hspace{8pt}
    \hfill
    \begin{subfigure}{0.23\textwidth}
        \centering
        \begin{tikzpicture}
            \begin{axis}[
                scale=0.5,
                xmode=log,             
                log basis x=2,
                xtick={2,4,8,16,32,64,128},
                xticklabels={2,4,8,16,32,64,128},
                yticklabel style={
                    font=\small,
                    /pgf/number format/fixed,
                    /pgf/number format/precision=2,
                    /pgf/number format/fixed zerofill
                },
                xlabel near ticks,
                % ylabel near ticks,
                title={Digg},
                xlabel={embedding dimension},
                grid=major,
                legend pos=south east,
                enlargelimits=0.15,
                legend style={font=\small, fill=none, inner sep=1pt}
            ]
            \addplot[
                color=blue!60,
                mark=*,
                mark size=1pt
                ]
                coordinates {
                (2, 0.8364) (4,0.7824) (8,0.8278) (16,0.8548) (32,0.8582) (64,0.8703) (128,0.8785)
                };
            \addlegendentry{1\%}
        
            \addplot[
                color=green,
                mark=*,
                mark size=1pt
                ]
                coordinates {
                (2, 0.7853) (4,0.6004) (8,0.8027) (16,0.8317) (32,0.8564) (64,0.8667) (128,0.8706)
                };
            \addlegendentry{5\%}
        
            \addplot[
                color=red,
                mark=*,
                mark size=1pt
                ]
                coordinates {
                (2, 0.7472) (4,0.8018) (8,0.7899) (16,0.8408) (32,0.8563) (64,0.8667) (128,0.8673)
                };
            \addlegendentry{10\%}
            \end{axis}
        \end{tikzpicture}
        % \caption{子图 2}
    \end{subfigure}
    \hfill
    % 第二行
    \begin{subfigure}{0.23\textwidth}
        \centering
        \begin{tikzpicture}
            \begin{axis}[
                scale=0.5,
                xmode=log,             
                log basis x=2,
                xtick={2,4,8,16,32,64,128},
                xticklabels={2,4,8,16,32,64,128},
                yticklabel style={
                    font=\small,
                    /pgf/number format/fixed,
                    /pgf/number format/precision=2,
                    /pgf/number format/fixed zerofill
                },
                xlabel near ticks,
                % ylabel near ticks,
                title={Bitcoin-Alpha},
                xlabel={embedding dimension},
                grid=major,
                legend pos=south east,
                enlargelimits=0.15,
                legend style={font=\small, fill=none, inner sep=1pt}
            ]
            \addplot[
                color=blue!60,
                mark=*,
                mark size=1pt
                ]
                coordinates {
                (2, 0.7502) (4,0.7315) (8,0.8706) (16,0.8797) (32,0.8815) (64,0.9404) (128,0.9728)
                };
            \addlegendentry{1\%}
        
            \addplot[
                color=green,
                mark=*,
                mark size=1pt
                ]
                coordinates {
                (2, 0.7088) (4,0.7660) (8,0.8287) (16,0.8505) (32,0.8949) (64,0.9217) (128,0.9607)
                };
            \addlegendentry{5\%}
        
            \addplot[
                color=red,
                mark=*,
                mark size=1pt
                ]
                coordinates {
                (2, 0.6760) (4,0.8468) (8,0.8556) (16,0.8797) (32,0.9047) (64,0.9263) (128,0.9398)
                };
            \addlegendentry{10\%}
            \end{axis}
        \end{tikzpicture}
        % \caption{子图 3}
    \end{subfigure}
    \hfill
    \begin{subfigure}{0.23\textwidth}
        \centering
        \begin{tikzpicture}
            \begin{axis}[
                scale=0.5,
                xmode=log,             
                log basis x=2,
                xtick={2,4,8,16,32,64,128},
                xticklabels={2,4,8,16,32,64,128},
                yticklabel style={
                    font=\small,
                    /pgf/number format/fixed,
                    /pgf/number format/precision=2,
                    /pgf/number format/fixed zerofill
                },
                xlabel near ticks,
                % ylabel near ticks,
                title={Bitcoin-OTC},
                xlabel={embedding dimension},
                grid=major,
                legend pos=south east,
                enlargelimits=0.15,
                legend style={font=\small, fill=none, inner sep=1pt}
            ]
            \addplot[
                color=blue!60,
                mark=*,
                mark size=1pt
                ]
                coordinates {
                (2, 0.8664) (4,0.9152) (8,0.8914) (16,0.9265) (32,0.9552) (64,0.9627) (128,0.9760)
                };
            \addlegendentry{1\%}
        
            \addplot[
                color=green,
                mark=*,
                mark size=1pt
                ]
                coordinates {
                (2, 0.7739) (4,0.9003) (8,0.8727) (16,0.9343) (32,0.9366) (64,0.9642) (128,0.9720)
                };
            \addlegendentry{5\%}
        
            \addplot[
                color=red,
                mark=*,
                mark size=1pt
                ]
                coordinates {
                (2, 0.7348) (4,0.8280) (8,0.9285) (16,0.9226) (32,0.9448) (64,0.9595) (128,0.9712)
                };
            \addlegendentry{10\%}
            \end{axis}
        \end{tikzpicture}
        % \caption{子图 4}
    \end{subfigure}

\caption{Sensitivity analysis results for the embedding dimension. Results are reported using the AUC-ROC score.}
    \label{embeddingdimension}
\end{figure*}
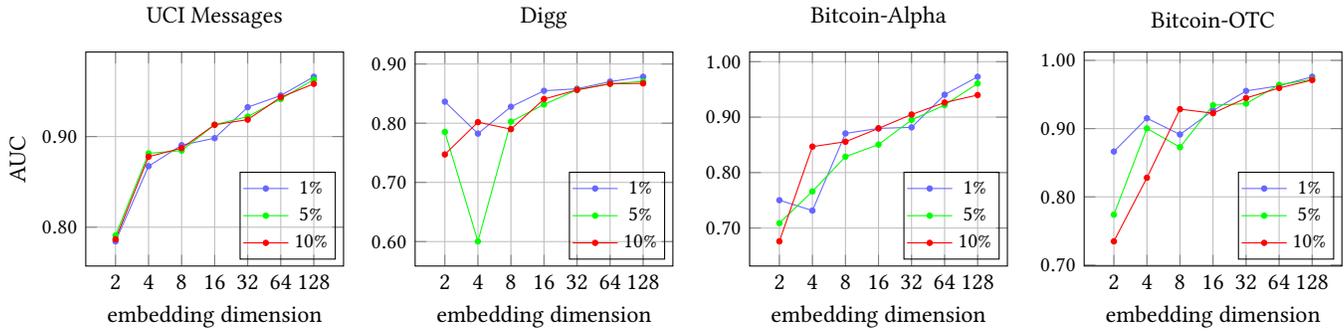

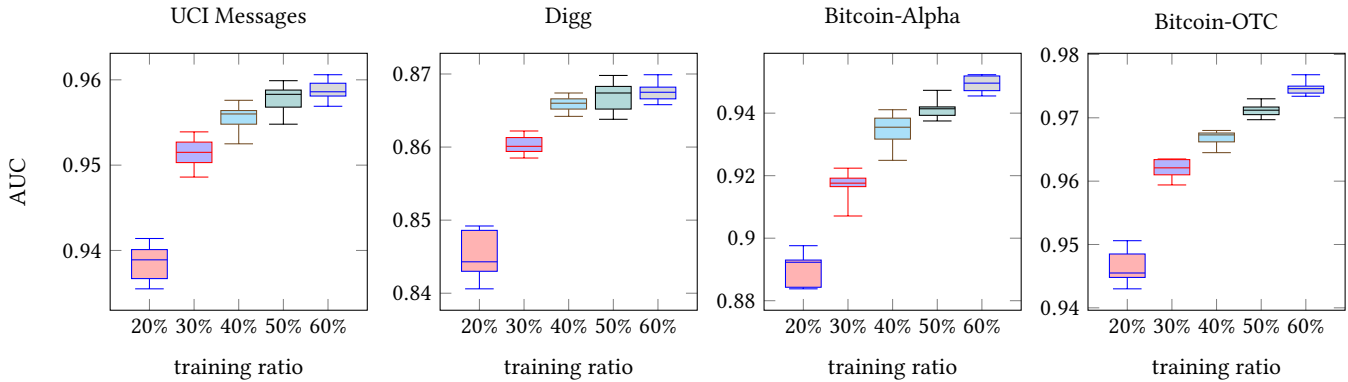
\begin{figure*}
    \centering
    \begin{subfigure}[b]{0.23\textwidth}
        \centering
        \begin{tikzpicture}
        \begin{axis}[
            boxplot/draw direction=y,
            ylabel={AUC},
            xlabel={training ratio},
            xtick={1,2,3,4,5},
            xticklabels={20\%,30\%,40\%,50\%,60\%},
            width=5cm,
            height=5cm,
            title={UCI Messages}
        ]

        \addplot+[
            boxplot prepared={
                median=0.9389,
                upper quartile=0.9401,
                lower quartile=0.9367,
                upper whisker=0.9414,
                lower whisker=0.9355,
            },
            fill=red!30
        ] coordinates {};

        \addplot+[
            boxplot prepared={
                median=0.9515,
                upper quartile=0.9527,
                lower quartile=0.9503,
                upper whisker=0.9539,
                lower whisker=0.9486,
            },
            fill=blue!30
        ] coordinates {};

        \addplot+[
            boxplot prepared={
                median=0.9560,
                upper quartile=0.9564,
                lower quartile=0.9548,
                upper whisker=0.9576,
                lower whisker=0.9525,
            },
            fill=cyan!30
        ] coordinates {};

        \addplot+[
            boxplot prepared={
                median=0.9583,
                upper quartile=0.9588,
                lower quartile=0.9568,
                upper whisker=0.9599,
                lower whisker=0.9548
            },
            fill=teal!30
        ] coordinates {};

        \addplot+[
            boxplot prepared={
                median=0.9586,
                upper quartile=0.9596,
                lower quartile=0.9581,
                upper whisker=0.9606,
                lower whisker=0.9569
            },
            fill=gray!30
        ] coordinates {};

        \end{axis}
        \end{tikzpicture}
    \end{subfigure}
    \hspace{20pt}
    \hfill
    % 可继续加第二个和第三个子图
    \begin{subfigure}[b]{0.23\textwidth}
        \centering
        \begin{tikzpicture}
        \begin{axis}[
            boxplot/draw direction=y,
            ylabel={},
            xlabel={training ratio},
            xtick={1,2,3,4,5},
            xticklabels={20\%,30\%,40\%,50\%,60\%},
            width=5cm,
            height=5cm,
            title={Digg}
        ]

        \addplot+[
            boxplot prepared={
                median=0.8443,
                upper quartile=0.8486,
                lower quartile=0.8430,
                upper whisker=0.8492,
                lower whisker=0.8406
            },
            fill=red!30
        ] coordinates {};

        \addplot+[
            boxplot prepared={
                median=0.8601,
                upper quartile=0.8613,
                lower quartile=0.8594,
                upper whisker=0.8622,
                lower whisker=0.8585
            },
            fill=blue!30
        ] coordinates {};

        \addplot+[
            boxplot prepared={
                median=0.8660,
                upper quartile=0.8666,
                lower quartile=0.8652,
                upper whisker=0.8674,
                lower whisker=0.8642
            },
            fill=cyan!30
        ] coordinates {};

        \addplot+[
            boxplot prepared={
                median=0.8674,
                upper quartile=0.8683,
                lower quartile=0.8652,
                upper whisker=0.8698,
                lower whisker=0.8638
            },
            fill=teal!30
        ] coordinates {};

        \addplot+[
            boxplot prepared={
                median=0.8675,
                upper quartile=0.8682,
                lower quartile=0.8666,
                upper whisker=0.8699,
                lower whisker=0.8658
            },
            fill=gray!30
        ] coordinates {};

        \end{axis}
        \end{tikzpicture}
    \end{subfigure}
    \hfill
    \begin{subfigure}[b]{0.23\textwidth}
        \centering
        \begin{tikzpicture}
        \begin{axis}[
            boxplot/draw direction=y,
            ylabel={},
            xlabel={training ratio},
            xtick={1,2,3,4,5},
            xticklabels={20\%,30\%,40\%,50\%,60\%},
            width=5cm,
            height=5cm,
            title={Bitcoin-Alpha}
        ]

        \addplot+[
            boxplot prepared={
                median=0.8923,
                upper quartile=0.8930,
                lower quartile=0.8843,
                upper whisker=0.8976,
                lower whisker=0.8838
            },
            fill=red!30
        ] coordinates {};

        \addplot+[
            boxplot prepared={
                median=0.9176,
                upper quartile=0.9192,
                lower quartile=0.9165,
                upper whisker=0.9224,
                lower whisker=0.9071
            },
            fill=blue!30
        ] coordinates {};

        \addplot+[
            boxplot prepared={
                median=0.9355,
                upper quartile=0.9384,
                lower quartile=0.9317,
                upper whisker=0.9411,
                lower whisker=0.9249
            },
            fill=cyan!30
        ] coordinates {};

        \addplot+[
            boxplot prepared={
                median=0.9414,
                upper quartile=0.9420,
                lower quartile=0.9393,
                upper whisker=0.9473,
                lower whisker=0.9375
            },
            fill=teal!30
        ] coordinates {};

        \addplot+[
            boxplot prepared={
                median=0.9496,
                upper quartile=0.9519,
                lower quartile=0.9472,
                upper whisker=0.9523,
                lower whisker=0.9455
            },
            fill=gray!30
        ] coordinates {};

        \end{axis}
        \end{tikzpicture}
    \end{subfigure}
    \hfill
    \begin{subfigure}[b]{0.23\textwidth}
        \centering
        \begin{tikzpicture}
        \begin{axis}[
            boxplot/draw direction=y,
            ylabel={},
            xlabel={training ratio},
            xtick={1,2,3,4,5},
            xticklabels={20\%,30\%,40\%,50\%,60\%},
            width=5cm,
            height=5cm,
            title={Bitcoin-OTC}
        ]

        \addplot+[
            boxplot prepared={
                median=0.9455,
                upper quartile=0.9485,
                lower quartile=0.9448,
                upper whisker=0.9506,
                lower whisker=0.9430
            },
            fill=red!30
        ] coordinates {};

        \addplot+[
            boxplot prepared={
                median=0.9621,
                upper quartile=0.9634,
                lower quartile=0.9610,
                upper whisker=0.9635,
                lower whisker=0.9594
            },
            fill=blue!30
        ] coordinates {};

        \addplot+[
            boxplot prepared={
                median=0.9673,
                upper quartile=0.9676,
                lower quartile=0.9662,
                upper whisker=0.9645,
                lower whisker=0.9680
            },
            fill=cyan!30
        ] coordinates {};

        \addplot+[
            boxplot prepared={
                median=0.9712,
                upper quartile=0.9717,
                lower quartile=0.9705,
                upper whisker=0.9730,
                lower whisker=0.9697
            },
            fill=teal!30
        ] coordinates {};

        \addplot+[
            boxplot prepared={
                median=0.9746,
                upper quartile=0.9750,
                lower quartile=0.9739,
                upper whisker=0.9768,
                lower whisker=0.9734
            },
            fill=gray!30
        ] coordinates {};

        \end{axis}
        \end{tikzpicture}
    \end{subfigure}
\caption{Sensitivity analysis results with different training ratios.}
    \label{trainingratio}
\end{figure*}

\subsection{Impact of Normality Distribution Shift}
To verify the impact of normality distribution shift on anomaly detection model performance, we design a controlled experiment by artificially introducing varying degrees of normality distribution shift into the datasets.

Specifically, since the original datasets do not contain raw node features, we first use node2vec to generate initial node features, denoted as $X^{0}$, and assign them as the node features for the graph snapshot $\mathcal{G}^{0}$ at timestamp $t = 0$.

For each subsequent timestamp $t > 0$, the node features $\boldsymbol{X^{t}}$ are generated by adding Gaussian noise with standard deviation $\sigma$ to the node features from the previous timestamp $\boldsymbol{X^{t-1}}$:
\begin{equation}
    \boldsymbol{X^t} = \boldsymbol{X^{t-1}} + \sigma\epsilon^t, \quad \epsilon^t \sim \mathcal{N}(0, I),
\end{equation}
where $I$ is the identity matrix, and $\epsilon^{t}$ is sampled independently for each node at each timestamp.

By introducing perturbations to the node features over time, we indirectly induce a shift in the edge distribution, thereby simulating the effect of NDS in dynamic graphs. After introducing the perturbations, we insert 10\% anomalous edges into each test set to evaluate anomaly detection performance under the shifted distributions. By adjusting $\sigma$, we control the strength of the injected distribution shift. We conduct experiments with $\sigma \in \{0.2, 0.4, 0.6, 0.8, 1.0\}$ to evaluate model robustness under different shift intensities.

We compare our proposed method with two baselines: RustGraph and our method without the NSEM module (w/o NSEM). For RustGraph~\cite{rustgraph2024}, we use their public code to re-run the experiment. The experimental results are shown in Figure~\ref{ndsexp}, from which we summarize the following findings:
\begin{itemize}
    \item Compared to RustGraph and the variant without NSEM, WhENDS consistently achieves the best performance across all experiments. As the intensity of NDS increases, WhENDS maintains relatively stable performance, and even under strong NDS, it continues to outperform other models, demonstrating its robustness to NDS.

    \item The NSEM module plays a critical role in mitigating the effects of NDS. Removing NSEM leads to significant performance degradation and a noticeable loss of stability under increasing NDS intensity, especially in UCI Messages and Bitcoin-Alpha.

    \item The model without NSEM still exhibits a certain level of robustness as the intensity of NDS increases. We attribute this to the numerically stable design of the Spatial-Temporal Encoder, which helps prevent excessive fluctuations in embedding values under large distribution shifts, thereby preserving some level of performance stability.
\end{itemize}

\subsection{Sensitivity Analysis}
We further conduct sensitivity analysis to investigate the impact of different hyperparameters on our model's performance.
Specifically, we evaluate the effects of varying the embedding dimension $d$ and the training ratio.
Experiments are conducted on all four datasets, and for each test set, we inject 10\% anomalous edges to evaluate model performance.

\vspace{1ex}
\noindent \textbf{Embedding Dimension $d$.} To investigate the effect of the embedding dimension, we vary $d$ across $\{2, 4, 8, 16, 32, 64, 128\}$ and evaluate model performance on all datasets.
As shown in Figure~\ref{embeddingdimension}, the model performance generally improves as the embedding dimension increases.
This indicates that a higher-dimensional embedding space allows the model to capture more informative representations, leading to better anomaly detection performance.

\vspace{1ex}
\noindent \textbf{Training Ratio.} To evaluate the effect of different training ratios, we conduct experiments with training ratios of \{20\%, 30\%, 40\%, 50\%, 60\%\}, and the results are summarized in Figure~\ref{trainingratio}.
The experimental results show that WhENDS benefits from higher training ratios, with its performance improving as more training data becomes available.
When the training ratio is low, WhENDS performs well on datasets with a larger number of snapshots, such as UCI Messages and Bitcoin-OTC, but struggles on Digg and Bitcoin-Alpha, where the number of snapshots is relatively limited.
We attribute this to the training strategy of the NSEM module. When there are sufficient snapshots, NSEM has access to a larger number of statistical samples across timestamps, which enables more effective learning of the statistics of normal edge embeddings.
In contrast, with fewer snapshots, the statistics available at each timestamp are limited, making it difficult to accurately estimate the statistics of normal edge embeddings, and leading to suboptimal performance of WhENDS.

\section{Conclusion}
In this paper, we identified normality distribution shift (NDS) as a fundamental challenge for anomaly detection in dynamic graphs, demonstrating that NDS can cause models to misclassify normal instances, thereby degrading performance and reducing robustness over time.  
To address this issue, we proposed \textbf{WhENDS}, a novel unsupervised anomaly detection method that mitigates the effects of NDS by aligning normal edge embeddings across timestamps to a standard Gaussian distribution via a whitening transformation.  
Extensive experiments demonstrate that WhENDS consistently outperforms strong baselines and remains robust under varying degrees of distribution shift.

Our model does not rely on any personally identifiable information or private attributes, and our experiments are conducted on publicly available benchmark datasets.

\bibliographystyle{plain}
\bibliography{refs}
\end{document}